\pdfoutput=1

\documentclass[11pt,a4paper]{article}
\usepackage[]{acl}

\usepackage{times}
\usepackage{latexsym}
\usepackage[bottom]{footmisc}

\usepackage[T1]{fontenc}

\usepackage[utf8]{inputenc}

\usepackage{graphicx}
\usepackage{booktabs}
\usepackage{adjustbox}
\usepackage{amsmath}
\usepackage{times}  
\usepackage{helvet}  
\usepackage{courier}  
\usepackage{algorithm}
\usepackage[noend]{algpseudocode}
\usepackage{tensor}
\usepackage{amsfonts}
\usepackage{float}
\usepackage{ragged2e}
\usepackage{tabularx}
\usepackage{hyperref}

\newcolumntype{b}{X}
\newcolumntype{s}{>{\hsize=.7\hsize}X}

\title{ MEKER: Memory Efficient Knowledge Embedding Representation for \\ Link Prediction and Question Answering }

\author{Viktoriia Chekalina\textsuperscript{\rm 1}, Anton Razzhigaev\textsuperscript{\rm 1,\rm 2}, Albert Sayapin\textsuperscript{\rm 1}, Evgeny Frolov\textsuperscript{\rm 1} \textbf{and Alexander Panchenko}\textsuperscript{\rm 1} \\
  \textsuperscript{\rm 1}Skolkovo Institute of Science and Technology, \textsuperscript{\rm 2}Artificial Intelligence Research Insitute (AIRI)}

\begin{document}

\maketitle

\begin{abstract}

Knowledge Graphs~(KGs) are symbolically structured storages of facts. The KG embedding contains concise data used in NLP tasks requiring implicit information about the real world. Furthermore, the size of KGs that may be useful in actual NLP assignments is enormous, and creating embedding over it has memory cost issues. We represent KG as a 3rd-order binary tensor and move beyond the standard CP decomposition~\cite{hitchcock-sum-1927} by using a data-specific generalized version of it~\cite{Hong_2020}. The generalization of the standard CP-ALS algorithm allows obtaining optimization gradients without a backpropagation mechanism. It reduces the memory needed in training while providing computational benefits. We propose a MEKER, a memory-efficient KG embedding model, which yields SOTA-comparable performance on link prediction tasks and KG-based Question Answering.

\end{abstract}

\begin{figure*}[htp]
\centering
\includegraphics[width=0.85\textwidth]{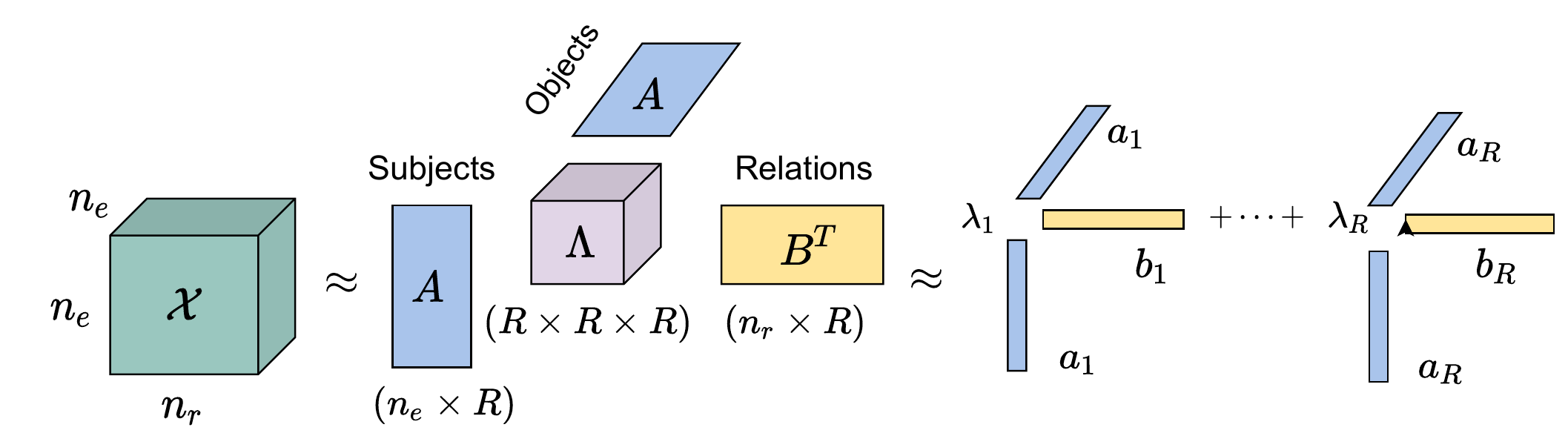} 
\caption{The CP decomposition scheme in the case of entity and relation KG embedding in MEKER. This is a binary 3-dimensional tensor $\mathcal{X}$ of knowledge graph facts that introduces objects, relations, and subjects indexes along the three axes. $B$ contains relation embedding, while $A$ represents entity vectors for the subject and object simultaneously. $\Lambda$ is the diagonal core tensor, identity in our case.}
\label{fig:cp_weights}
\end{figure*}

\section{Introduction}

Natural Language Processing~(NLP) models have taken a big step forward over the past few years. For instance, language models can generate fluent human-like text without any problems. However, some applications like question answering and recommendation systems need correct, precise, and trustworthy answers.

For this goal, it is appropriate to leverage knowledge graphs~(KG)~\cite{freebase, YAGO2016} a structured repository of essential facts about the real world. 
For convenience, the knowledge graph can be represented as a set of triples. A triple is two entities bound with relation and describes the fact. It takes the forms of $\langle e_s, r, e_o \rangle$, where $e_s$ and $e_o$ represent objects and subject entities, respectively. 





For efficient use of information from KG, there is a need for the low-dimensional embedding of graph entities and relations. KG embedding models usually use a standard Neural Networks~(NN) backward mechanism for parameter tuning, duplicating its memory consumption. Hence, existing approaches to embedding learning have substantial memory requirements and can be deployed only on small datasets under a single GPU card. Processing large KGs appropriate for the custom downstream task is a challenge.

There are several libraries designed to solve this problem. 
Framework LibKGE~\cite{ruffinelli2020you} allows the processing of large datasets by using sparse embedding layers. Despite the memory saving, sparse embedding has several limitations - for example, in the PyTorch library, they are not compatible with several optimizers.
PyTorch-BigGraph~\cite{pbg} operates with large knowledge graphs by dividing them into partitions - distributed subgraphs. Subgraphs need a place for storing, embedding models need modifications to work with partitions and perform poorly.

The main contribution of our paper is a memory-efficient approach to learning Knowledge Graph embeddings MEKER~(Memory Efficient Knowledge Embedding Representation). It allows more efficient KG embedding learning, maintaining comparable performance to state-of-the-art models. MEKER leverages generalized canonical Polyadic~(CP) decomposition~\cite{Hong_2020}, which allows a better approximation of given data and analytical computation of the parameters' gradient. MEKER is evaluated on a link prediction task using several standard datasets and large datasets based on Wikidata. Experiments show that MEKER achieves highly competitive results on these two tasks. To demonstrate downstream usability, we create a Knowledge Base Question Answering system Text2Graph and use embeddings in it. The system with MEKER embeddings performs better as compared to other KG embeddings, such as PTBG~\cite{pbg}.



\section{Related Work}

There are three types of approaches for learning KG embedding: distance-based, tensor-based, and deep learning-based models. The first group is based on the assumption of translation invariance in the embedding vector space. In model \textbf{TransE}~\cite{TranE} relations are represented as connection vectors between entity representations. \textbf{TransH}~\cite{TransH} implies relation as a hyperplane onto which entities are being projected.
\textbf{QuatE}~\cite{QuatE} extends the idea with hypercomplex space and represents entities as embeddings with four imaginary components and relations as rotations in the space.

Tensor-based models usually represent triples as a binary tensor and look for embedding matrices as factorization products. \textbf{RESCAL}~\cite{Nickel_11} employs tensor factorization in the manner of DEDICOM~\cite{Harshman_DEDIDCOM}, which decomposes each tensor slice along the relationship axis. \textbf{DistMult}~\cite{Yang2015EmbeddingEA} adapts this approach by restricting the relation embedding matrix to diagonal. On the one hand, it reduces the number of relation parameters, on the other hand, it losses the possibility of describing asymmetric relations. The \textbf{ComplEX}~\cite{Complex} represents the object and subject variants of a single entity as complex conjugates vectors. It combines tensor-based and translation-based approaches and solves the asymmetric problem. \textbf{TuckER}~\cite{balazevic-etal-2019-tucker} uses Tucker decomposition~\cite{Tuck1966c} for finding representation of a knowledge graph elements. This work can also be considered a generalization of several previous link prediction methods. 

Standard Canonical Polyadic~(CP)~\cite{hitchcock-sum-1927} decomposition in the link prediction task does not show outstanding performance~\cite{KG_via_CTF}. Several papers address this problem by improving the CP decomposition approach. \textbf{SimplIE}~\cite{Simplie} states that low performance is due to different representations of subject and object entity and deploys CP decomposition with dependently learning of subjects and objects matrices. \textbf{CP-N3}~\cite{CP_N3} highlights the statement that the Frobenius norm regularizing is not fit for tensors of order more than 3~\cite{cheng16} and proposes a Nuclear p-norm instead of it. Our approach also uses CP decomposition with enhancement. We consider remark from~\textbf{SimplIE} and set the object and subject representations of one entity to be equals. At the same time, inside the local step of the CP decomposition algorithm, the matrices of subjects and objects consist of different elements and are different (see Appendix). In contradistinction to \textbf{CP-N3}, we do not employ a regularizer to improve training but change the objective. Instead of squared error, we use logistic loss, which is appropriate for one-hot data. We abandon the gradient calculation through the computational graph and count gradient analytically, which makes the training process less resource-demanding.

Approaches based on Deep Learning convolutions and attention mechanisms \textbf{ConvE, GAT, GAAT}~\cite{ConvE, GAT, GAAT} achieve high performance in link prediction. Besides, they have their disadvantages - it necessitate more time and memory resources than other types of models and usually needs pre-training.

\section{MEKER: Memory Efficient Knowledge Embedding Representation}

Our approach to entity embeddings relies on generalized CP tensor decomposition~\cite{hitchcock-sum-1927}. Namely, $R$-rank CP decomposition approximates an N-dimensional tensor as a sum of $R$ outer products of $N$ vectors. Every product can also be viewed as a rank-$1$ tensor. This approximation is described by the following formula: $\mathcal{X} \approx \mathcal{M} = [| A, B, C|]$, where $\mathcal{X} \in \mathbb{R}^{I\times J \times K}$ is original data and $\mathcal{M} \in \mathbb{R}^{I\times J \times K}$ is its approximation. Factors have the following shape $A \in \mathbb{R}^{I\times R}$, $B \in \mathbb{R}^{J\times R}$, $C \in \mathbb{R}^{K\times R}$. The scheme of CP decomposition applied to the KG elements representation task is in Figure~\ref{fig:cp_weights}. We set matrix $A$ equal to matrix $C$ and simultaneously corresponding to subject and object entities. 


\subsection{Generalization of Canonical Poliyadic (CP) Decomposition}

Following the determination of the approximation type, the next task is to find the parameters of the factor matrices that best match the ground truth data. \newcite{CP_ALS_Kolda, WCP_ALS_Kolda} describe the most widely used CP decomposition algorithm, CP-ALS. The update rules for the factor matrices are derived by alternating between minimizing squared error~(MSE) loss.~\newcite{Hong_2020} demonstrates that MSE corresponds to Gaussian data and is a particular case of a more general solution for an exponential family of distributions. In general, the construction of optimal factors originates the minimization problem:

\begin{equation}
\begin{split}
    \min F (\mathcal{M}; \mathcal{X}) & \equiv \sum_{i \in \Omega} f(x_i, m_i), \\
    f(x, m) & \equiv  \log p (x | l^{-1}(m)), \\
\end{split}
\label{eq:2}
\end{equation}
where $f$ - elementwise loss function, $\Omega$ - set of indices of known elements of $\mathcal{X}$, $l$ - link function, $x_i$ and $m_i$ - the $i$-th elements of $\mathcal{X}$ and $\mathcal{M}$, respectively. We also introduce $\mathcal{Y}$ - the tensor of derivatives of the elementwise loss with the same size as $\mathcal{X}$ and being filled by zeros for $i \not\in \Omega$. The data in the sparse one-hot triple tensor has a Bernoulli distribution. The link function for Bernoulli is $l(\rho) = log(\rho / (1 - \rho))$ and associated probability is $\rho = \exp(m)(1 - \exp(m))$ so the loss function and elements of the $\mathcal{Y}$ are defines as follows:
\begin{equation}
\begin{aligned}
f(x_i, m_i) &= \log(1 + \exp{m_i}) - x_i m_i, \\
y(x_i, m_i) &= \frac{\partial{f(x_i, m_i)}}{\partial{m_i}} = \frac{\exp{m_i}}{1 + \exp{m_i}} - x_i.\\
\end{aligned}
\label{eq:3}
\end{equation}

\newcite{Hong_2020} derives partial derivatives of $F$ w.r.t. factor matrices and presents gradients G of it in a form similar to standard CP matrix update formulas:

\begin{equation}
\begin{aligned}
     G_A &= \mathcal{Y}_{[0]}(B\odot C)^{T\dagger}, \\
     G_B &= \mathcal{Y}_{[1]}(A\odot C)^{T\dagger}, \\
     G_C &= \mathcal{Y}_{[2]}(A\odot B)^{T\dagger}, \\
\end{aligned}
\label{eq:2}
\end{equation}
 where $\dagger$ - pseudo-inverse matrix, $\odot$ - Khartri-Rao operator, $\mathcal{X_{[\text{n}]}}$ - mode-n matricization, a reshaping of tensor $\mathcal{X}$ along the $n$ axis.
The importance of representation~(\ref{eq:2}) is that we can calculate the gradients via an essential tensor operation called the matricized tensor times Khatri-Rao product~(MTTKRP), implemented and optimized in most programming languages.
Algorithm~\ref{alg:gcp_grad} describes the procedure for computing factor matrices gradients~(\ref{eq:2}) in a Bernoulli distribution case~(\ref{eq:3}).

\subsection{Implementation Details}

We use PyTorch~\cite{pytorch} to implement the MEKER model. We set the object and subject factors equal and correspond to matrix A for the decomposition of the one-hot KG triplet tensor. Sparse natural and reconstructed tensors are stored in Coordinate Format as a set of triplets~(COO). We combine actual triples and sampled negative examples in batches, and process them. The corresponding pieces from the ground-truth tensor and current factor matrices are cut out for each batch. Then the pieces are sent to Algorithm~\ref{alg:gcp_grad} for the calculation of gradients of the matrix elements with appropriate indexes. Algorithm~\ref{alg:gcp_wrap} describes the pseudocode of factorization KG tensor using GCP gradients.

We train the MEKER model using Bayesian search optimization to obtain the optimal training parameters.
We use the Wandb.ai tool~\cite{wandb} for experiment tracking and visualizations. The complete sets of tunable hyperparameters are in the Appendix. Table~\ref{table:greed_search} shows the best combinations of it for the proposed datasets. 

\subsection{Baselines}
As a comparison, we deploy related link prediction approaches that meet the following criteria: 1) it should learn KG embedding from scratch 2) it should report high performance 3) the corresponding paper should provide a runnable code. We use the Tucker, Hyper, ConvKB, and QuatE implementations from their respective repositories. For TransE, DistMult, ComplEx, and ConvE, we use LibKGE~\cite{ruffinelli2020you} library with the best parameter setting for reproducing every model. We run each model five times for each observed value and provide means and sample standard deviation. 

\begin{algorithm}[H]
\fontsize{9.0pt}{9.0pt}\selectfont
\caption{GCP GRAD Bernuilli}\label{alg:gcp_grad}
\textbf{Input}: $\mathcal{X}$ \Comment{Ground Truth Tensor}\\
                $A$, $B$, $C$ \Comment{Factor matrices} \\
                
\textbf{Output}: $F$, $G_A$, $G_B$, $G_C$\\

\begin{algorithmic}
\State $\mathcal{M} = \{A, B, C\}$ \Comment{Model Restored tensor}\\
\State $F = \sum_i f(x_i, m_i) = \sum log(1 + e^m_i) - x_i m_i$ \Comment{Loss}\\
\State $\mathcal{Y} = \sum_i \frac{\delta f(x_i, m_i)}{\delta m_i} =$ \Comment{Derivative tensor}\\
$= \sum \frac{1}{1 + e^{(-m_i)}} - x_i$ \\
\State $G_A = \mathcal{Y}_{[0]}(B\odot C)^{T\dagger}$.              \Comment{Element-wise gradient for A}
\State $G_B = \mathcal{Y}_{[1]}(A \odot C)^{T\dagger}$              \Comment{Element-wise gradient for B}       
\State $G_C = \mathcal{Y}_{[2]}(A \odot B)^{T\dagger}$              \Comment{Element-wise gradient for C}
\end{algorithmic}
\end{algorithm}

\begin{algorithm}[H]
\fontsize{9pt}{9pt}\selectfont
\caption{Factorization of the KG tensor using GCP gradients}\label{alg:gcp_wrap}

\textbf{Input}: $\mathcal{X}$ \Comment{Ground Truth Tensor}\\
                Triplets \Comment{List of triplets }\\
                $LR$ \Comment{learning rate}\\
                $R$ \Comment{Desired size of embeddings}\\
                $N$ \Comment{Number of epoch}\\

\textbf{Output}: $A$, $B$ \Comment{Updated factor matrices}\\

Initialize factor matrices $A \in \mathbb{R}^{R\times n_e}$, $B \in \mathbb{R}^{R\times n_r}$ \\
\begin{algorithmic}
\For{$i = 1 \dots N$}
\For {$[\text{inds}_a, \text{inds}_b, \text{inds}_c]$ \textbf{in} Triplets}
    \State $\mathcal{X}_{batch} = \mathcal{X}[\text{inds}_a,\text{inds}_b,\text{inds}_c]$
    
    \State $g_a$, $g_b$, $g_c$, $loss$ = \\
    GCP\_GRAD$(\mathcal{X}_{batch}, A[\text{inds}_a]$, $B[\text{inds}_b]$, $A[\text{inds}_c]$)
    
    \State $A[\text{inds}_a]\text{.grad} = g_a$
    \State $B[\text{inds}_b]\text{.grad} = g_b$
    \State $A[\text{inds}_c]\text{.grad} = g_c$
    \State UPDATE($A$, $B$, $LR$)
\EndFor

\EndFor
\end{algorithmic}
\end{algorithm}

\begin{table*}[ht]
\fontsize{9pt}{9pt}\selectfont%
\centering 
\begin{tabular}{@{}p{5.3cm}| >{\RaggedLeft}p{0.85cm} >{\RaggedLeft}p{0.85cm} >{\RaggedLeft}p{0.85cm} >{\RaggedLeft}p{0.85cm} >{\RaggedLeft}p{0.85cm} >{\RaggedLeft}p{0.85cm} >{\RaggedLeft}p{0.85cm} >{\RaggedLeft}p{0.85cm}@{}} 
\hline
Dataset &  &  & FB15k237 &  &  & WNRR18 &  &  \\
\midrule
Model & MRR & Hit@10 & Hit@3 & Hit@1 & MRR & Hit@10 & Hit@3 & Hit@1  \\ 

\midrule
ConvKB~\cite{ConvKB} & 0.2985 & 0.4785 & 0.3270 & 0.2296 & 0.2221 & 0.5074 & 0.3777 & 0.0347 \\
HypER~\cite{Hyper}& 0.3423 & 0.5228 & 0.3774 & 0.2536  & 0.4653 & 0.5228 & 0.4774 & 0.4361 \\
TuckER~\cite{balazevic-etal-2019-tucker} & 0.3455 & \underline{0.5408} & 0.3899 & 0.2606  & 0.4654 & 0.5215 & 0.4784 & 0.4368 \\
QuatE~\cite{QuatE} &  \textbf{0.3614} & \textbf{0.5538} & \textbf{0.4014} & \textbf{0.2711} & \textbf{0.4823} & \textbf{0.5719} & \textbf{0.4955} & \textbf{0.4360} \\
CP-N3~\cite{CP_N3}&  0.3514 & 0.5294 & 0.3876 & 0.2646 & 0.4402 & 0.4858 & 0.4485 & 0.4207 \\
\midrule 
LibKGE ConvE~\cite{ConvE} & 0.3367 & 0.5213 & 0.3682 & 0.2381  & 0.4282 & 0.5049 & 0.4492 & 0.3934 \\
LibKGE TransE~\cite{TranE}  & 0.3121 & 0.4962 & 0.3175 & 0.2195 & 0.2274 & 0.5189 &  0.3677 & 0.0516 \\
LibKGE DistMult~\cite{Yang2015EmbeddingEA} & 0.3331 & 0.5185 & 0.3673 & 0.2410 & 0.4505 & 0.5215 & 0.4634 & 0.4162\\
LibKGE ComplEx~\cite{Complex} & 0.3390 & 0.5265 & 0.3724 & 0.2468 & 0.4752 & \underline{0.5467} & 0.4809 & 0.4366\\

\midrule
MEKER & \underline{0.3588} & 0.5393 & \underline{0.3915} & \underline{0.2682}  & \underline{0.4768} & 0.5447 & \underline{0.4875} & \underline{0.4371} \\
\bottomrule
\end{tabular}
\caption{Link Prediction scores for various models on the FB15k237 and WN18RR datasets. The embedding size is 200. The winner scores are highlighted in bold font, and the second results are underlined.}
\label{table:fb15237_new} 
\end{table*}

\begin{table}[ht]
\fontsize{10.0pt}{10.0pt}\selectfont%
\centering 
\begin{tabular}{@{}lrr@{}} 
\toprule 
Dataset  & FB15k237 & WN18RR\\
\midrule
Optimizer & AdamW &  AdamW \\
LR & 0.01 &  0.009   \\
Batch Size  & 156 &  128 \\
L2 reg & 0.001 &  0.0 \\
Number of negative & 6 &  8 \\
Step of decay LR & 3 &  15 \\
Gamma of decay LR & 0.8 &  0.6 \\
\bottomrule
\end{tabular}
\caption{The best hyperparameters of the MEKER.} 
\label{table:greed_search} 
\end{table}

\section{Experiments on Standard Link Prediction Datasets}
\subsection{Experimental settings}
The Link prediction task estimates the quality of KG embedding. Link prediction is a classification predicting if triple over graph elements is true or not. The scoring function $\Phi(e_s, rel, e_o)$ returns the probability of constructing a true triple. 
 We test our model on this task using standard Link prediction datasets. 

\textbf{FB15k237}~\cite{toutanova-chen-2015-observed} is a dataset based on the FB15k237 adapted Freebase subset, which contains triples with the most mentioned entities. FB15k237 devised the method of selecting the most frequent relations and then filtering inversions from test and valid parts. The \textbf{WN18RR}~\cite{TranE} version of WN18 is devoid of inverse relations. WN18 is a WordNet database that contains the senses of words as well as the lexical relationships between them. Table~\ref{table:dataset_statistic} shows the number of entities, relations, and train-valid-test partitions for each dataset used in the proposed work. As an evaluation, we obtain complementary candidates from the entity set for each pair entity-relation from each test triple and estimate the probability score of the received triple being true. The presence of a rising real supplement entity at the top indicates a hit. Candidate ranking is provided using a filtered setting, which was first used in~\cite{TranE}. In a filtered setting, all candidates who completed a true triple on the current step are removed from the set, except for the expected entity. We use $\text{Hit@}1$, $\text{Hit@}3$, $\text{Hit@}10$ as evaluation metrics. We also use mean reciprocal rank (MRR) to ensure that true complementary elements are ranked correctly.

\begin{table}[ht]
\fontsize{8.0pt}{8.0pt}\selectfont%
\centering 
\begin{tabular}{@{}p{1.3cm}|>{\RaggedLeft}p{1.1cm}|>{\RaggedLeft}p{0.65cm}|>{\RaggedLeft}p{1.15cm}>{\raggedleft\arraybackslash}p{0.75cm}>{\raggedleft\arraybackslash}p{0.75cm}@{}} 
\toprule 
& & & \multicolumn{3}{c}{Number of Triplets} \\
Dataset  & \#ents & \#rels & Train & Valid & Test  \\ 
\midrule
Fb15k237 & 14,541 & 237 & 27.2$\cdot10^4$ & 17,535 & 20,466\\
WN18RR & 40,943 & 11 & 8.6$\cdot10^4$ & 30,034 &  3,134 \\
Wiki4M & 4,316$\cdot 10^4$ & 1,245& 1,367$\cdot10^4$& 30,000 &  35,815 \\
Wikidata5m & 4,594$\cdot 10^4$ & 822 & 2,061$\cdot 10^4$ & 5,163 & 5,133 \\
\bottomrule
\end{tabular}
\caption{Statistics of link prediction datasets.} 
\label{table:dataset_statistic} 
\end{table}


\subsection{Link Prediction}
Table~\ref{table:fb15237_new} shows the mean value of the experiment on small datasets for the embedding of size 200. The $\text{Hit@}10$ standard deviation for MEKER is $0.0034$ for the FB15k237 dataset and $0.0026$ for the WNRR18 dataset. Due to space constraints, the table with deviations from all experiments, comparable to Table~\ref{table:fb15237_new}, is in Appendix.

The best score belongs to QuatE~\cite{QuatE} model due to its highly expressive 4-dimensional representations. Among the remaining approaches, MEKER outperforms its contestants' overall metrics except for the Hit@10 - Tucker model surpasses MEKER for Fb15k237, ComplEX by LibKGE for WNRR18.
In general, MEKER shows decent results comparable to strong baselines~\cite{QuatE, balazevic-etal-2019-tucker}. It is also worth noting that MEKER significantly improves MRR and Hit@1 metrics on freebase datasets, whereas on word sense, according to data, it has been enhanced in Hit@10.

\subsection{Model efficiency in case of parameter size increasing}


With a strong memory assumption, we can reduce the size of pre-trained MEKER embeddings by tenfold while losing only a few percent of performance.

\begin{figure*}[h]
  \centering
  
  \begin{minipage}[b]{0.43\textwidth}
   \centering
    \includegraphics[width=.99\linewidth]{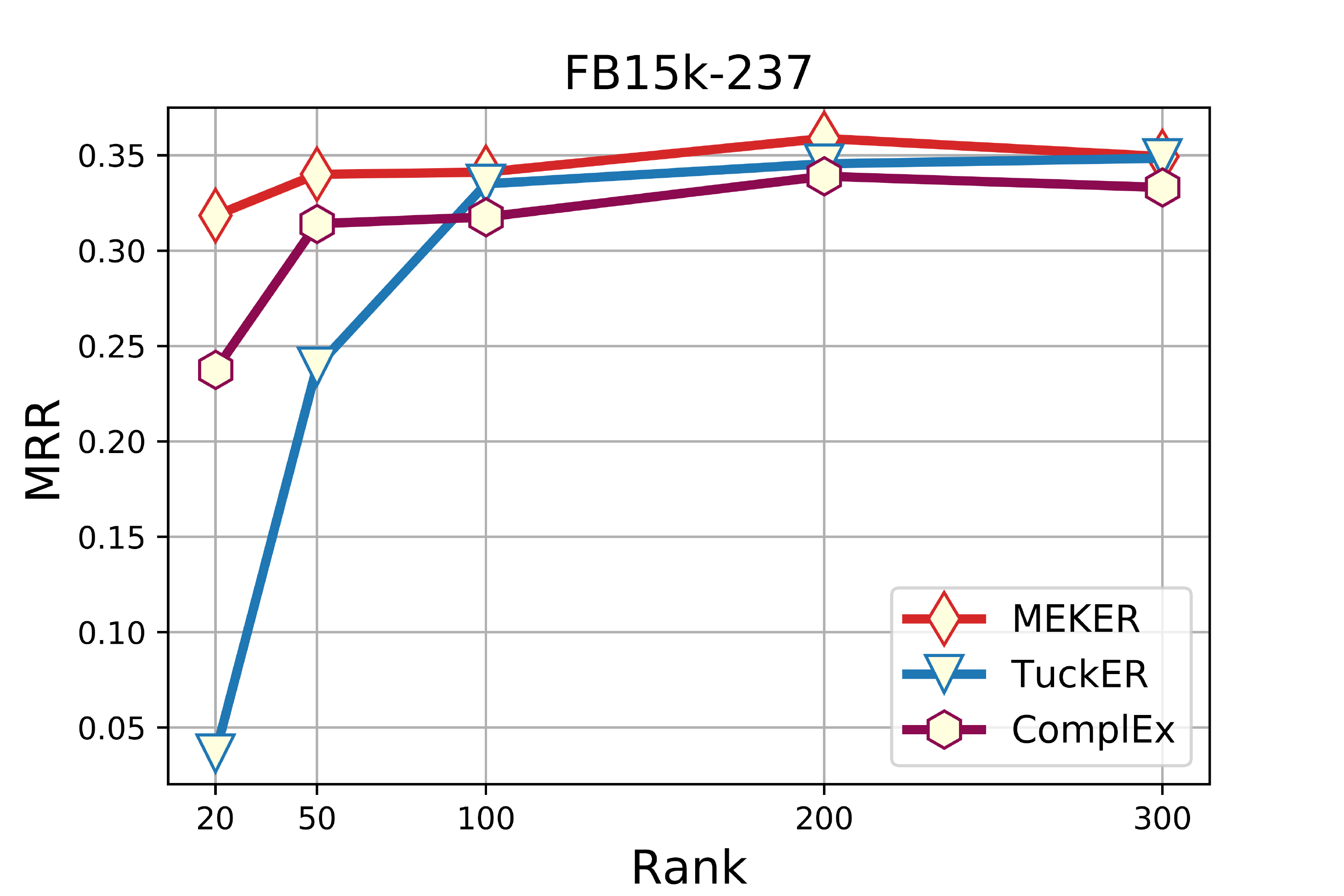}
    \caption{MRR score in dependence of embedding ranks}
    \label{fig:ranks0}
  \end{minipage}
  \hspace{0.11\textwidth}
  \centering
  \begin{minipage}[b]{0.43\textwidth}
  \centering
    \includegraphics[width=0.99\linewidth]{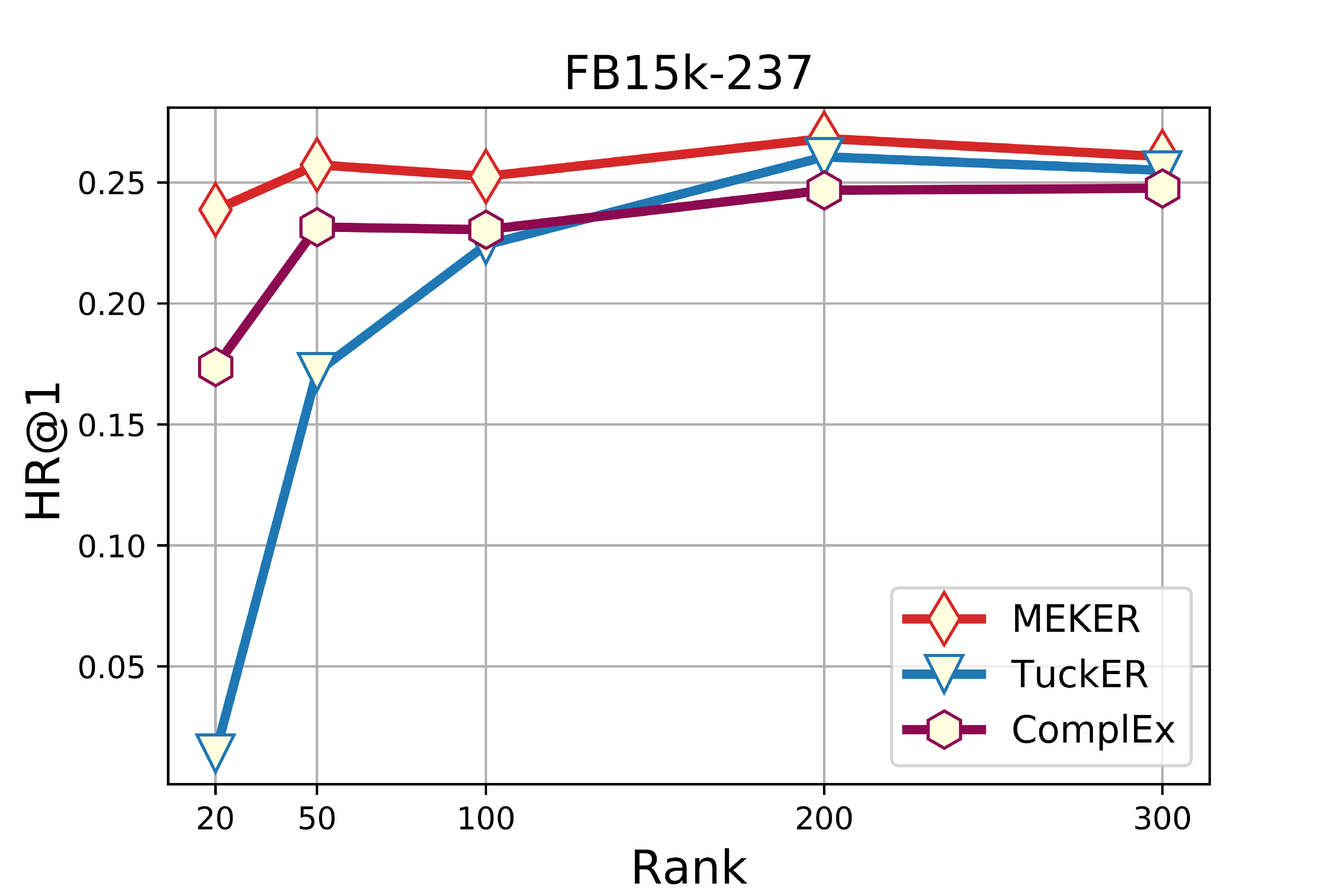}
    \caption{Hit@1 score in dependence of embedding ranks}
    \label{fig:ranks1}
  \end{minipage}
\label{fig:ranks}
\end{figure*}

Figures~\ref{fig:ranks0},~\ref{fig:ranks1} show MRR and Hit@1 scores for MEKER, TuckER, and ComplEX models at various embedding sizes. Each model approaches a constant value on both metrics around rank 100. For ranks 200 and 300, the performance difference between the three models is approximately consistent for both metrics, with MEKER scoring the highest on rank 20. It means that the number of MEKER parameters can be reduced while maintaining or improving quality. The quality loss is significant for other presented models.
\subsection{Memory Complexity Analysis}



\begin{figure}[htp]
  \centering
\includegraphics[width=.89\linewidth]{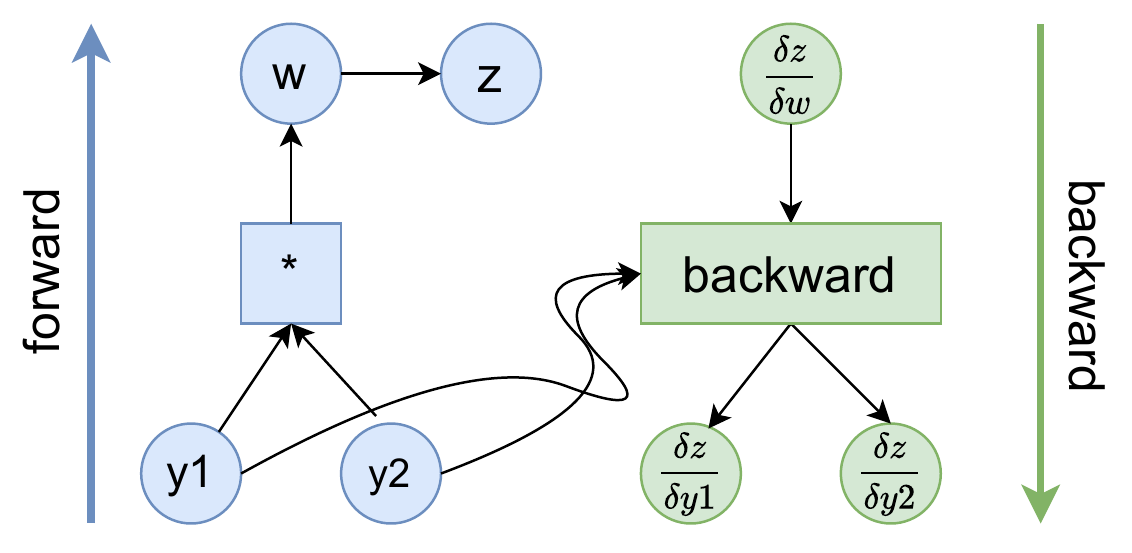}
\caption{The scheme of the augmented computational graph of the Neural Network.}
\label{fig:comp_graphs}
\end{figure}

The theoretical space complexity of models mentioned in the current work is shown in the right column of Table~\ref{table:memory}. In the context of the Link Prediction task, all approaches have asymptotic memory complexity $\mathcal{O}((n_e + n_r)d)$, which is proportional to the size of the full dictionary of KG elements, i.e. the embedding layer or look-up table. Other aspects of the proposed models are less significant: the convolutional layers are not very extensive. The implementation determines the amount of real memory used by the model during the training process. The Neural Network backpropagation mechanism is used to tune parameters in the most related work. Backpropagation in Figure~\ref{fig:comp_graphs} creates computational graph in which all model parameters are duplicated. It results in a multiplicative constant $2$, insignificant in a small dictionary but becomes critical in a large one. To summarize, the following factors account for the decrease in MEKER's required memory:
\begin{enumerate}
    \item In the MEKER algorithm gradients are computed analytically. 
    \item MEKER does not have additional neural network layers (linear, convolutional, or attention). 
\end{enumerate}

To measure GPU RAM usage, we run each considered embedding model on FB15k-237 into a single GPU and print peak GPU memory usage within the created process. The left column of a Table~\ref{table:memory} demonstrates that MEKER has objective memory complexity that is at least twice lower than that of other linear approaches. This property reveals the possibility of obtaining representations of specific large databases using a single GPU card.

\begin{table}[ht]
\fontsize{9pt}{8.5pt}\selectfont%
\centering 
\begin{tabular}[!h]{@{}lrr@{}} 
\toprule 
Model  & GPU Memory & Theoretical Approximation  \\ 
 &  Usage, MB & of Space Complexity \\
\midrule 

TuckER  & 357 & $2\cdot((n_e + n_r + c\cdot lin)\cdot d)$ \\

HypER  & 208 & $ 2\cdot((n_e + n_r + c\cdot lin)\cdot d)$ \\
ConvKB  & 3\,563 & $ 2\cdot((n_e + n_r)\cdot d + c\cdot conv)$\\
ConvE  & 229 & $ 2\cdot((n_e + n_r)\cdot d + c\cdot conv)$\\
ComplEX  & 252 & $ 2\cdot(n_e + n_r)\cdot d$\\
DistMult  & 174 & $2\cdot(n_e + n_r)\cdot d$\\
QuatE & 2\,367 & $2\cdot 4 \cdot (n_e + n_d + c\cdot lin) $\\ 
CP~(N3) & 138 & $2\cdot(n_e + n_r)\cdot d$ \\
\midrule
MEKER & 79  & $((n_e + n_r)\cdot d)$\\ 
\bottomrule
\end{tabular}
\caption{Memory, reserved in the PyTorch Framework during the training process and theoretical approximation of given implementations' complexity. On the FB15k237 dataset, we train 200-size representations with a batch size of 128. $Lin$ denotes the number of output features in a linear layer, $conv$ denotes the size of convolutional layer parameters. The constant $c$ represents the number of different layers.} 
\label{table:memory} 
\end{table}

\section{Experiments on Large-Scale KG Datasets}
\subsection{Experimental settings}

To test the model on large KG, we employ two WikiData-based datasets. The first English dataset, \textbf{Wikidata5m}~\cite{wang2021KEPLER}\footnote{\url{https://deepgraphlearning.github.io/project/wikidata5m}}, is selected due to the presence of related works and reproducible baseline~\cite{ruffinelli2020you}. This dataset is created over the 2019 dump of WikiData and contains of elements with links to informative Wikipedia pages. Our experiments use the transductive setting of Wikidata5m - triplet sets to disjoint across training, validation, and test. 

The second English-Russian dataset is formed since its suitability for the NLP downstream task. We leverage KG-based fact retrieval over Russian Knowledge Base Questions~(RuBQ)~\cite{rybin2021rubq} benchmark. This benchmark is a subset of Wikidata entities with Russian labels. Some elements in RuBQ are not covered with Wikidata5m, so we created a link-prediction \textbf{Wiki4M} dataset over RuBQ. We select triples without literal objects and obtain approximately 13M triples across 4M entities~(see Table~\ref{table:dataset_statistic}). Wiki4M also fits the concept of multilingualism is intended to be used in a cross-lingual transfer or few-shot learning.


\begin{table*}[h]
\fontsize{9.0pt}{9.0pt}\selectfont%
\centering 
\begin{tabular}{@{}lrrrrrr@{} } 
\toprule 
Model  & MRR & Hit@1 & Hit@3 & Hit@10 &  Memory, GB & Storage, GB  \\ 
\midrule 

\multicolumn{7}{c}{\textit{English: Wikidata5m dataset}} \\ \midrule 

PTBG~(ComplEX) & 0.184  & 0.131 & 0.210 & 0.287 & 45.15  & 9.25 \\

PTBG~(TransE)  & 0.150  & 0.091 & 0.176 & 0.263 & 43.64  & 9.25\\

LibKGE sparse~(TransE) & 0.142  & 0.153 & 0.211 & 0.252 & 33.29  & 0.00\\

LibKGE sparse ~(ComplEX) & 0.202 & \textbf{0.160} & 0.233 & 0.316 & \textbf{21.42}  & 0.00 \\

MEKER~(ours) & \textbf{0.211} & 0.149 &\textbf{0.238}  & \textbf{0.325} & 22.27 & 0.00 \\

\midrule
\multicolumn{7}{c}{\textit{Russian: Wiki4M dataset}} \\ \midrule 

PTBG~(ComplEX) & 0.194  & 0.141 & 0.212 & 0.293 & 42.83  & 9.25 \\
LibKGE sparse~(TransE)  & 0.183  & 0.126 & 0.191 & 0.275 & 26.75  & 0.00\\

LibKGE sparse ~(ComplEX)  & 0.247  & 0.196 & 0.275 & 0.345 & \textbf{20.22}  & 0.00\\

MEKER~(ours) & \textbf{0.269} & \textbf{0.199} & \textbf{0.303} & \textbf{0.410} & 21.04 & 0.00 \\

\bottomrule
\end{tabular}
\caption{Unfiltered link prediction scores for MEKER and PyTorch-BigGraph approaches for Wiki4M and Wikidata5m datasets and memory needed in leveraging every model. Storage means additional memory demanded for auxiliary structures. Batch size 256. Here ``RAM'' is GPU RAM or main memory RAM if GPU limit of 24 GB is reached. \textit{Sparse} means sparse embeddings. Models without \textit{sparse} mark employ dense embeddings matrix.}
\label{table:wiki4m} 
\end{table*}






\subsection{Link Prediction}

We embed the datasets for ten epochs on a 24.268 Gb GPU card with the following model settings: LR $2.5 \cdot 10^{-4}$, increasing in $0.5$ steps every $10$ epoch, batch size $256$, number of negative samples $4$ for Wiki4M and $2$ for Wikidata5m.

As a comparison, we use the PyTorch-BigGraph large-scale embedding system~\cite{pbg}. PyTorch-BigGraph modifies several traditional embedding systems to focus on the effective representation of KG in memory. We select ComplEX and TransE and train graphs for these embedding models, dividing large datasets into four partitions. With a batch size of $256$, the training process takes $50$ epochs.

We also deploy LibKGE~\cite{ruffinelli2020you} to evaluate TransE and ComplEX approaches. For ComplEX model training, we use the best parameter configuration from the repository, for TransE, we obtain a set of training parameters by greed search. The learning rate for TransE is $0.5$, decaying in factor $0.45$ every $5$ step and train model in $100$ epochs. In both cases, we use sparse embedding in the corresponding model setting and batch size of $256$. Models from both wrappers that did not fit in 24 GB, we train on the CPU.

Embedding sets yielded by we these experiments we then test on the link prediction task. We provide scoring without filters because the partition-based setup of PyTorch-Biggraph does not support filtering evaluation. Tables~\ref{table:wiki4m} shows that MEKER significantly improves the results of PyTorch-Biggraph models across all proposed metrics. The ComplEX model with sparse embedding, fine-tuned by LibKGE, gives results almost approaching the MEKER and exceeding the Hit@1 in Wiki4M. The right part of Tables~\ref{table:wiki4m} shows that the baseline approaches consume twice as much memory as MEKER, but sparse ComplEX slightly improves memory consumption. TransE does not give such significant results as ComplEX.

\subsection{Knowledge Base Question Answering~(KBQA)}

In this section, to further evaluate the proposed MEKER embeddings we test them in an extrinsic way within on a KBQA task on two datasets for English and Russian.

\subsubsection{Experimental Setting}


We perform experiment with two datasets: for English we use the common dataset SimpleQuestions~\cite{SimpleQuestions} aligned with Wiki4M KG\footnote{\url{https://github.com/askplatypus/wikidata-simplequestions}} (cf. Table~\ref{table:dataset_statistic}), and for Russian we use RuBQ 2.0 dataset~\cite{rybin2021rubq} which comes with the mentioned above Wiki4M KG (cf. Table~\ref{table:dataset_statistic}). RuBQ 2.0 is a Russian language QA benchmark with multiple types of questions aligned with Wikidata. For both SimpleQuestions and RuBQ, for each question, an answer is represented by a KG triple. 

For training we use a training set of SimpleQuestions for verification we use a test set of SimpleQuestions and RuBQ 2.0 dataset for English and Russian, respectively. These Q\&A pairs provide ground truth answers linked to exact this version of KG elements.

More specifically, in these experiments, we test answers to 1-hop questions which are questions corresponding to one subject and one relation in the knowledge graph, and takes their object as an answer. 

We want to leverage the KBQA model, which can process questions both in English and Russian. To measure the performance of a KBQA system, we measure the accuracy of the retrieved answer/entity. This metric was used in previously reported results on SimpleQuestions and RuBQ. If the subject of the answer triple matches the reference by ID or name, it is considered correct.

\begin{figure*}[htp]
\centering
\includegraphics[width=0.8\textwidth]{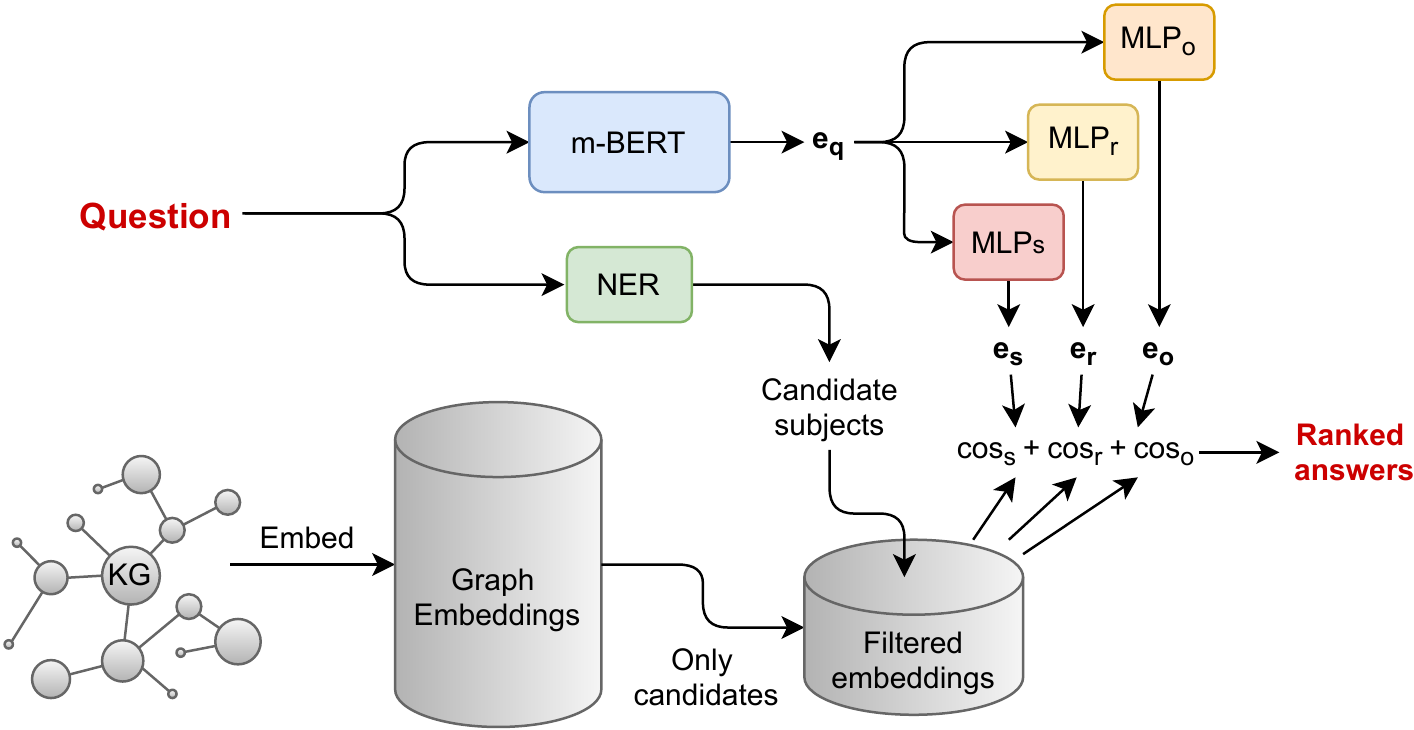}
\caption{Text2Graph method used in our experiments: 1-Hop QA pipeline. First, we take original entity and relation embeddings. The question is embedded using m-BERT. This embedding is then processed by MLP, yielding candidate representations of an object, relation, and subject. The sum of the subject, relation, and object cosines is the final score of triple candidates.}
\label{fig:1hopQA}
\end{figure*}

\subsubsection{KBQA methods}

The key idea of the KBQA approaches is mapping questions in natural language to the low-dimensional space and comparing them to graph elements' given representation. In \textbf{KEQA}~\cite{KBQA} LSTM models detect the entity and predicates from the question text and project it further into the entity and predicate embedding spaces. The closest subject in terms of similarity to the entity and predicate embeddings is selected as the answer.

We created a simple approach~\textbf{Text2Graph} which stems from the \textbf{KEQA} and differs from the original work in improved question encoder, entity extractor, additional subject embedding space and simplified retrieval pipeline.
The Algorithm~\ref{alg:text2graph} describes the procedure of projecting the input question to graph elements. The multilingual-BERT~\cite{bert} model encodes the input question, and all word vectors are averaged into a single deep contextualized representation $e_q$. This representation then goes through three MLPs jointly learning candidate embeddings of an object, relation, and subject. We minimize MSE between predicted embeddings and the corresponding KGE model's embeddings. The appropriateness score of every fact in KG is a sum of cosine similarity between MLP outputs and ground truth model representation for every element in the triple. The triple with the highest score is considered to be an answer. The scheme is trained using an AdamW optimizer with default parameters for $10$ epochs.

\subsection{Baselines}

\subsubsection{RuBQ 2.0} 
 We compare our method to several QA approaches compatible with questions from this benchmark. 

\begin{algorithm}[H]
\fontsize{9.5pt}{9.5pt}\selectfont
\caption{Text2Graph question projection algorithm}
\label{alg:text2graph}
\textbf{Input}: $\mathcal{Q}$, $\mathcal{G}$, \textbf{E},\\ text encoder $M_{enc}$,\\ projection modules: $M_s, M_r, M_o$,\\
Subject Candidates Extractor: \text{NER}\\
\textbf{Output}: answer $\langle  o_a, r_a, s_a\rangle$
\begin{algorithmic} 
\State $e_q = M_{enc}(\mathcal{Q})$
\State Initialize answers-candidates list with empty list \textbf{A}=[]
\State Initialize scores list with empty list \textbf{S}=[]
\State Initialize entities-candidates list with empty list \textbf{C}=[]
\For{entity \textbf{in} $\mathcal{G}$}
    \If{entity.name \textbf{in} $\text{NER}(\mathcal{Q})$}
    \State \textbf{C}.append(entity)
    \EndIf
\EndFor
\For{entity \textbf{in} \textbf{C}}
    \For{relation \textbf{in} entity.relations}
        \State s = entity.id
        \State r = relation.id
        \State o = entity[r]
        
        \State triple = $\langle  s, r, o\rangle$
        \State \textbf{A}.append(triple)
        \State $\mathbf{e_s} = \textbf{E}${\text[s]}
        \State $\mathbf{e_r} = \textbf{E}${\text[r]}
        \State $\mathbf{e_o} = \textbf{E}${\text[o]}
        \State $\mathbf{y_s} = M_s(e_q)$
        \State $\mathbf{y_r} = M_r(e_q)$
        \State $\mathbf{y_o} = M_o(e_q)$
        \State $score = \text{cos}(\mathbf{e_o},\mathbf{y_o}) + \text{cos}(\mathbf{e_r},\mathbf{y_r}) + \text{cos}(\mathbf{e_s},\mathbf{y_s})$
        \State \textbf{S}.append($score$)
\EndFor
\EndFor
\State ind = argmax(\textbf{S})
\State $\langle  s_a, r_a, o_a\rangle = \textbf{A}$[ind]
\State \textbf{return} $\langle  s_a, r_a, o_a\rangle$ 
\end{algorithmic}
\end{algorithm}

\textbf{QAnswer}\footnote{\url{https://www.qanswer.eu}} is a rule-based system addressing questions in several languages, including Russian. \textbf{SimBa} is a baseline presented by RuBQ 2.0 authors. It is a SPARQL query generator based on an entity linker and a rule-based relation extractor. KBQA module of \textbf{DeepPavlov Dialogue System Library}~\cite{burtsev-etal-2018-deeppavlov} also based on query processing.

\subsubsection{SimpleQuestions}

\textbf{Simple Question} is an English language benchmark aligned with FB5M KG - the subset of Freebase KG. Its train and validation parts consist of 100k and 20k questions, respectively.
As a baseline solution we employ \textbf{KEQA}~\cite{KBQA}. We realign answers from this benchmark to our system, which is compatible with Wikidata5m. Not all of the questions from FB5M have answers among Wiki4M, that is why we test both systems on a subset of questions whose answers are present in both knowledge graphs.

\subsubsection{Experimental Results}

\begin{table}[tb]
\fontsize{9.5pt}{9.5pt}\selectfont%
\centering 
\resizebox{0.48\textwidth}{!}{
\begin{tabular}{@{}lrr@{}} 
\toprule 
KBQA Model & Embedding Model & Accuracy 1-Hop\\ 
\midrule 
DeepPavlov & -  & 30.5 $\pm$ 0.04 \\
SimBa & -  & 32.3 $\pm$ 0.05 \\
QA-En & -  & 32.3 $\pm$ 0.08 \\
QA-Ru & -  & 30.8 $\pm$ 0.03 \\
\midrule 
Text2Graph & PTBG~(ComplEX) Wiki4M & 48.16 $\pm$ 0.05 \\
Text2Graph & PTBG~(TransE) Wiki4M & 48.84 $\pm$ 0.06\\
Text2Graph& MEKER Wiki4M & 49.06 $\pm$ 0.06\\
\bottomrule
\end{tabular}
}
\caption{Comparison of the Text2Graph system with the various KG embeddings with existing solutions (QA-Ru, QA-En, SimBa) on RuBQ 2.0 benchmark.}
\label{table:rubg_acc} 
\end{table}

\begin{table}[tb]
\fontsize{9.5pt}{9.5pt}\selectfont%
\centering 
\resizebox{0.48\textwidth}{!}{
\begin{tabular}{@{}lrr@{}} 
\toprule 
KBQA Model & Embedding Model & Accuracy 1-Hop\\ 
\midrule 
KEQA & TransE FB5M & 40.48 $\pm$ 0.10\\
\midrule 
Text2Graph & PTBG~(TransE) Wikidata5m & 59.97 $\pm$ 0.15\\
Text2Graph& MEKER Wikidata5m & 61.81 $\pm$ 0.13\\
\bottomrule
\end{tabular}
}
\caption{Comparison of the Text2Graph system with the various KG embeddings with existing embedding-based solution on the SimpleQuestions benchmark.} 
\label{table:sq_acc} 
\end{table}

We compare the results of the Text2Graph with PTBG embeddings versus MEKER embedding and baseline KBQA models. 
Results on the RuBQ 2.0 dataset are shown in Table~\ref{table:rubg_acc}. Text2Graph outperforms baselines. Using MEKER embeddings instead of the PTBG version of ComplEX and TransE demonstrates slightly better accuracy.

Table~\ref{table:sq_acc} presents results on the SimpleQuestions dataset. As \citet{KBQA} model uses FB5M KG and Text2Graph uses Wikidata5m KG we test both models on the subset of questions, which answers are present in both knowledge graphs for a fair comparison. Our model demonstrates superior performance and regarding the comparison within different embeddings in a fixed system, MEKER provides better accuracy of answers than TransE embeddings on the SimpleQuestions benchmark.


\section{Conclusion}

We propose MEKER, a linear knowledge embedding model based on generalized CP decomposition. This method allows for the calculation of gradient analytically, simplifying the training process under memory restriction. In comparison to previous KG embedding linear models~\cite{balazevic-etal-2019-tucker}, our approach achieves high efficiency while using less memory during training. On the standard link prediction datasets WN18RR and FB15k-237, MEKER shows quite competitive results. 

In addition, we created a Text2Graph — KBQA system based on the learned KB embeddings to demonstrate the model's effectiveness in NLP tasks. We obtained the required representations using MEKER on the Wikipedia-based dataset Wiki4M for questions in Russian and on Wikidata5m for questions in English. Text2Graph outperforms baselines for English and Russian, while using MEKER's  embeddings provides additional performance gain compared to PTBG embeddings. Furthermore, our model's link prediction scores on Wiki4M and Wikidata5m  outperform the baseline results. MEKER can be helpful in question-answering systems over specific KG, in other words, in systems that need to embed large sets of facts with acceptable quality.

All codes to reproduce our experiments are available online.\footnote{\url{https://github.com/skoltech-nlp/meker}}

\section*{Acknowledgements}

The work was supported by the Analytical center under the RF Government (subsidy agreement 000000D730321P5Q0002, Grant No. 70-2021-00145 02.11.2021).

We thank Dr.~Evgeny Frolov for providing advice and assistance in developing the MEKER method.

\bibliography{custom}
\bibliographystyle{acl_natbib}

\newpage
\clearpage
\appendix

\end{document}